%% file: main.tex
\documentclass{article}

\usepackage{microtype}
\usepackage{graphicx}
\usepackage{subfigure}
\usepackage{booktabs} 

\usepackage{hyperref}

\usepackage[accepted]{icml_fmwild}

\usepackage{amsmath}
\usepackage{amssymb}
\usepackage{mathtools}
\usepackage{amsthm}

\usepackage[capitalize,noabbrev]{cleveref}

\theoremstyle{plain}

\theoremstyle{definition}

\theoremstyle{remark}

\usepackage[textsize=tiny]{todonotes}

\usepackage{booktabs}
\usepackage{multirow}
\usepackage{colortbl}
\usepackage[OT1]{fontenc} 
\usepackage{threeparttable} 

\begin{document}

\twocolumn[
\icmltitle{
VFA: Vision Frequency Analysis of Foundation Models and Human
}




\begin{icmlauthorlist}
\icmlauthor{Mohammad-Javad Darvishi-Bayazi}{mila,udem}
\icmlauthor{Md Rifat Arefin}{mila,udem}
\icmlauthor{Jocelyn Faubert}{udem}
\icmlauthor{Irina Rish}{mila,udem}
\end{icmlauthorlist}

\icmlaffiliation{mila}{Mila, Québec AI Institute, Montréal, QC, Canada}
\icmlaffiliation{udem}{Université de Montréal, Montréal, QC, Canada}

\icmlcorrespondingauthor{Mohammad-Javad Darvishi-Bayazi}{mj.darvishi92@gmail.com}

\icmlkeywords{Machine Learning, ICML}

\vskip 0.3in
]



\printAffiliationsAndNotice{}  

\begin{abstract}

Machine learning models often struggle with distribution shifts in real-world scenarios, whereas humans exhibit robust adaptation. 
Models that better align with human perception may achieve higher out-of-distribution generalization.
In this study, we investigate how various characteristics of large-scale computer vision models influence their alignment with human capabilities and robustness. Our findings indicate that increasing model and data size and incorporating rich semantic information and multiple modalities enhance models' alignment with human perception and their overall robustness. Our empirical analysis demonstrates a strong correlation between out-of-distribution accuracy and human alignment.

\end{abstract}

\section{Introduction}

The deployment of machine learning models in real-world scenarios is challenging due to distribution shifts \cite{koh2021wilds}. Several methods have attempted to improve models' out-of-distribution (OOD) generalization by learning robust representations \cite{gulrajani2020search}. Humans, on the other hand, exhibit remarkable robustness to distribution shifts. It is argued that aligning models with human perception can enhance their robustness \cite{geirhos2018imagenettrained, fel2022harmonizing}.

\begin{figure}[tbhp]
  \centering
  \includegraphics[width=0.45\textwidth]{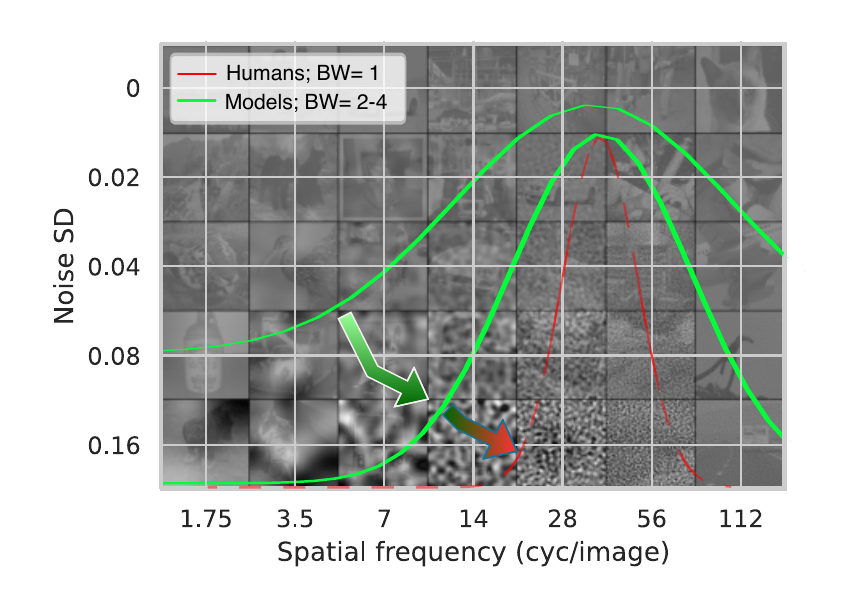}
    \caption{Frequency Bandwidths of Humans and Models. Humans are sensitive to a narrow frequency band, and adding noise within this band (under the red curve) degrades their performance. In contrast, models exhibit a wider frequency band (green curves), making them more vulnerable to noise across a broader range of frequencies. Narrowing the band might improve robustness.}
      \label{fig:abs}
\end{figure}

To compare these two systems, we need a method to assess not only their performance but also their underlying mechanisms. Frequency analysis is a promising approach to studying human vision \cite{campbell1968application}. By masking a specific frequency band, we can analyze a system's sensitivity to those frequencies and identify the most critical band for tasks such as object recognition. 
Recently, critical frequency band masking has been used to study Artificial Neural Networks (ANNs) \cite{subramanian2024spatial}. It has been shown that \textbf{humans} recognize objects in natural images using a narrow, one-octave-wide channel, which is consistent across various stimuli such as letters and gratings \cite{solomon1994visual, majaj2002role}, establishing it as a canonical feature of human object recognition. In contrast, \textbf{ANNs} utilize frequency channels that are $2-4$ times wider than those of humans (see Figure \ref{fig:abs}), making them sensitive to a broader range of frequencies \cite{subramanian2024spatial} and therefore prone to failure in real-world applications.

In this study, we conduct an extensive exploration of numerous computer vision models to answer these questions:
    1) \emph{Are ANNs similar to humans in object recognition tasks?}
    2) \emph{Can modern computer vision models match or outperform humans amid frequency noise?}
    3) \emph{What factors contribute to their proximity to human performance?}
    4) \emph{Humans rely more on the shape of objects than their texture, and models are texture-biased \cite{geirhos2018imagenet}. Therefore, can Bandwidth (BW) predict shape bias?}
    5) \emph{Would decreasing the bandwidth to be closer to that of humans improve the robustness?}

\begin{figure*}[tbh] 

  \centering
  \includegraphics[width=0.75\textwidth]{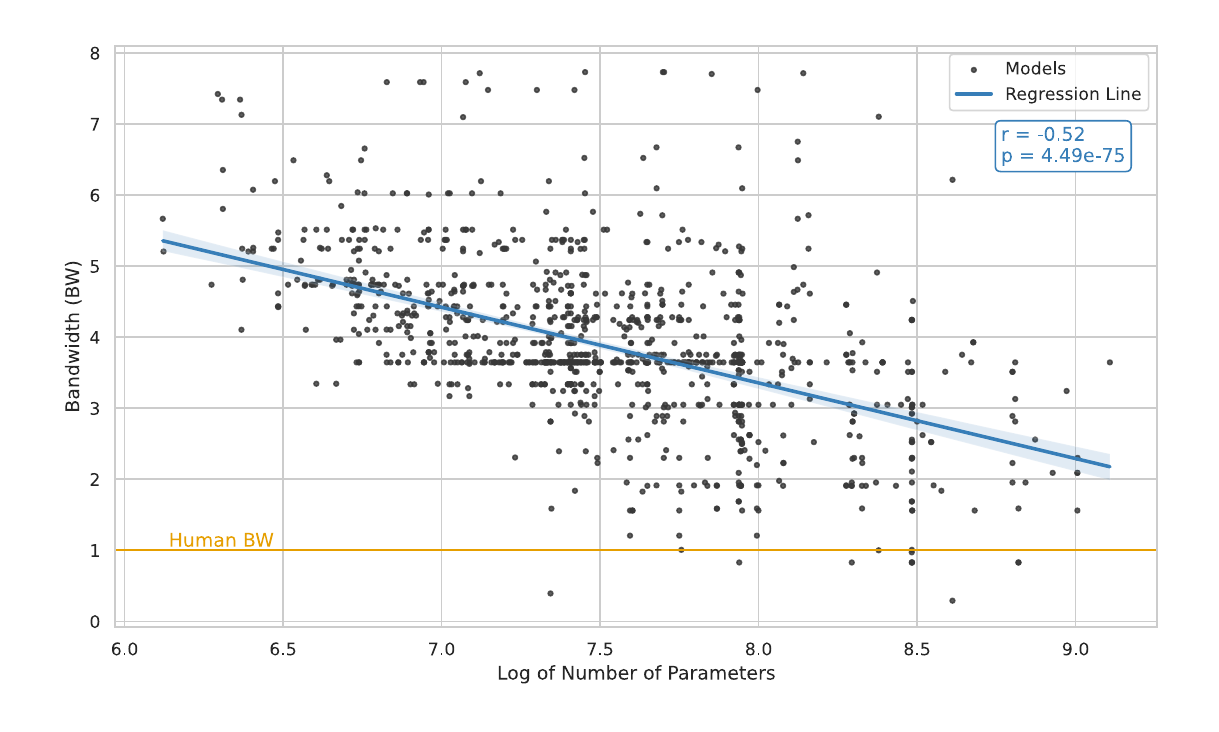}
\caption{Correlation between Bandwidth (BW) and Model Size in Logarithmic Scale. The regression line represents that as the model size scales, the bandwidth decreases, converging towards human levels. Each dot corresponds to a model, for model names and details, see Section \ref{sec:Scaling_Experiment_details} in the Appendix.}
  \label{fig:model_scale}
\end{figure*}

\begin{figure}[tbh]
  \centering
  \includegraphics[width=0.45\textwidth]{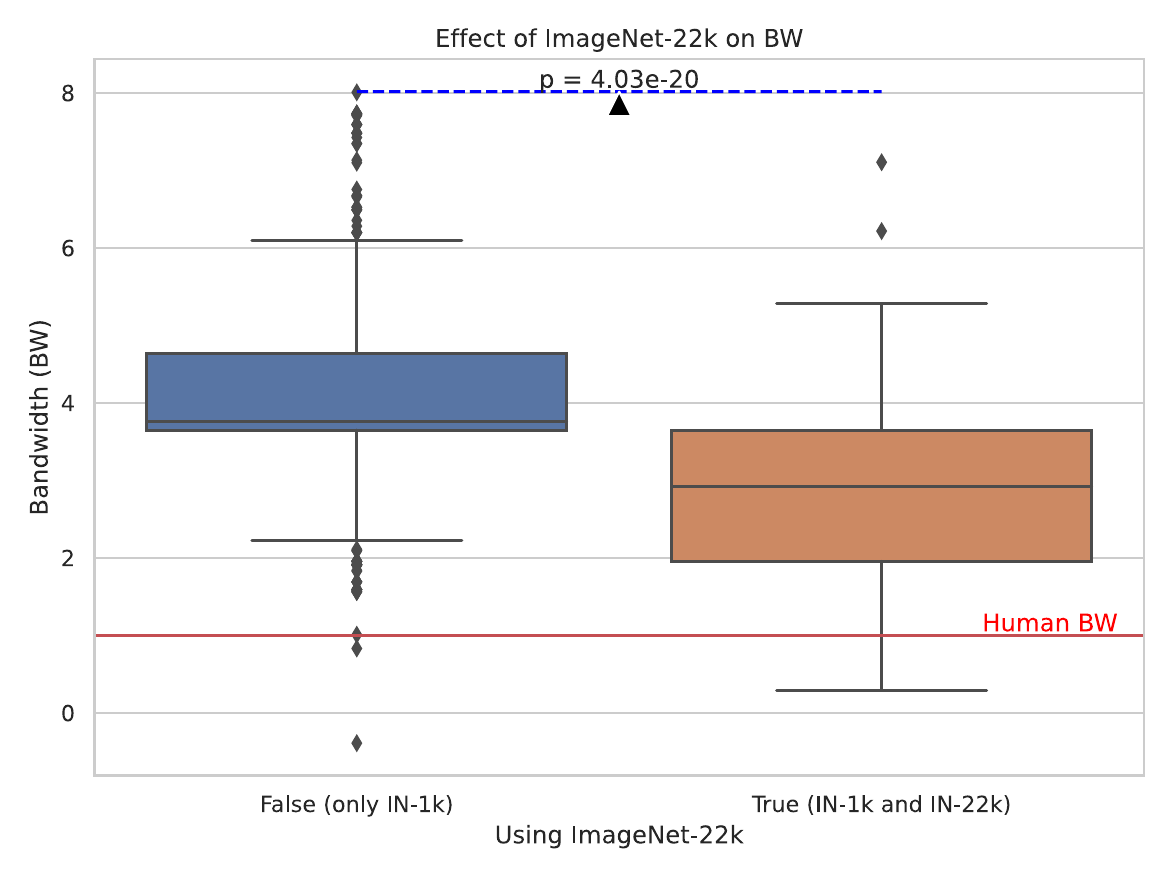}
\caption{Effect of Data on Bandwidth: Comparing models trained with ImageNet-22K. Models that benefit from ImageNet-22K training exhibit significantly smaller bandwidths.}
  \label{fig:data_scale}

\end{figure}

\section{Related work}

Over the years, different methods have been proposed to address poor generalization under distribution shifts ~\citep{zhang2022towards, du2020metanorm, li2018learning, sun2016deep, sagawa2019distributionally, shi2021gradient}. However, the underlying principles for better generalization remain unknown. Comparison of the robustness of models to humans has also been studied, as humans show more robustness to distribution changes~\citep{geirhos2021partial}. Inspired by human robustness, ~\citet{fel2022harmonizing} propose a strategy to align models with human behavior. Recently, based on frequency analysis\cite{campbell1968application}, critical frequency band masking has been applied to models \cite{subramanian2024spatial}. Humans recognize objects in natural images using a narrow one-octave-wide channel, consistent with stimuli such as letters and gratings \cite{solomon1994visual, majaj2002role}, establishing it as a canonical feature of human object recognition. Our study is inspired by this frequency analysis to understand the behavior of the models and their robustness analysis based on different metrics such as OOD accuracy and shape bias as introduced in \citep{geirhos2021partial}. 

\section{Methodology}
We explore what characteristics of ANNs can close their gap with human performance. We follow the same procedure as \cite{subramanian2024spatial}, adding different spatial noise in various frequency bands and different noise standard deviations (SD) as shown in Figure \ref{fig:abs}. Then we evaluate the systems using these distorted images and fit a Gaussian curve to the point where they reach the $50\%$ accuracy threshold. We calculate the bandwidth as the logarithm of the width at half-maximum in octaves. 
In this work, we tested \textbf{more than 1200} discriminative models available on HuggingFace timm \cite{rw2019timm}, multimodal zero-shot and fine-tuned CLIP \cite{pmlr-v139-radford21a,ilharco_gabriel_2021_5143773, cherti2023reproducible} models to analyze their bandwidth and robustness.

\begin{figure*}[htbp!]  
    \centering
    \includegraphics[width=0.75\textwidth]{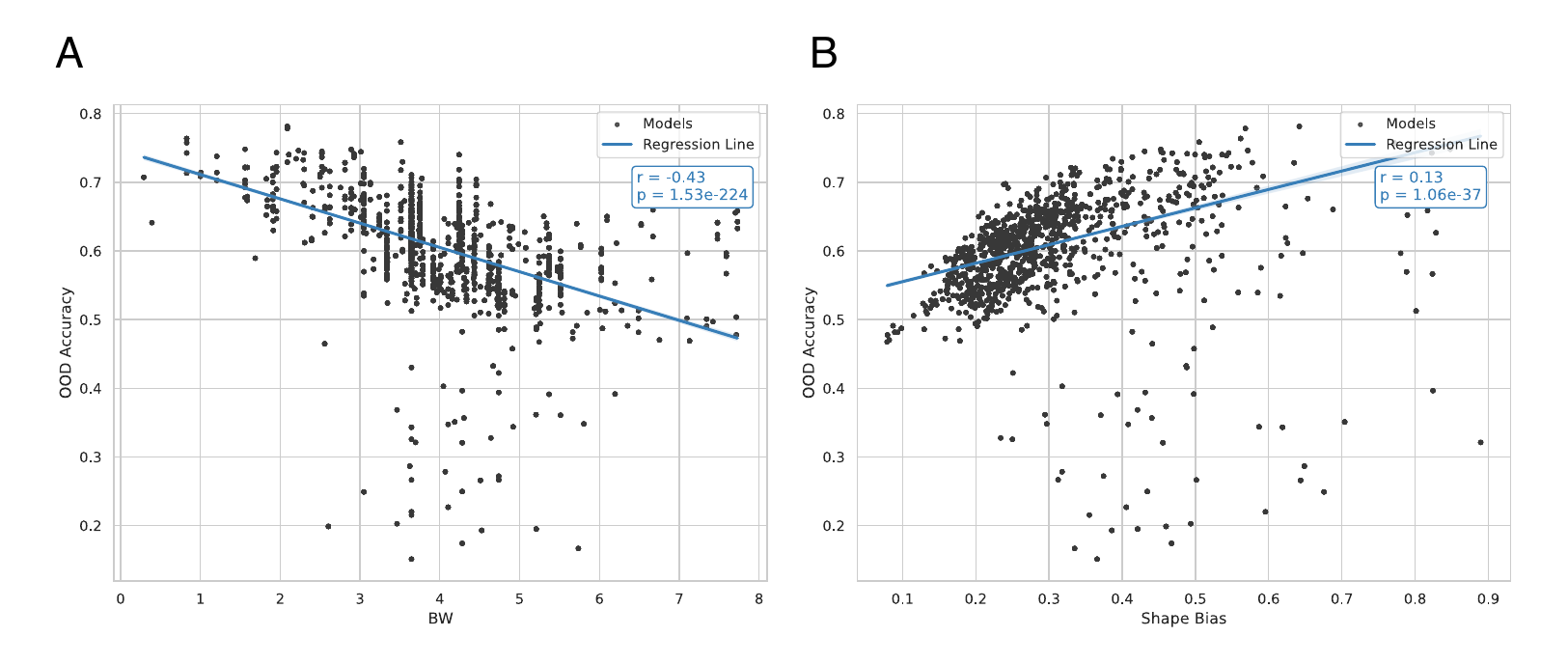}
    \caption{BW and Shape bias comparison. A) OOD versus Bandwidth. B) OOD versus Shape bias. Both BW and Shape bias are predictive of OOD performance and are correlated, with BW showing a higher correlation with OOD generalization.}
    \label{fig:BW_ShB}
\end{figure*}
To examine the OOD accuracy and shape bias of models and humans, we utilized a comprehensive collection of 17 OOD datasets curated by \citet{geirhos2021partial}. These datasets contain 16 superclasses of ImageNet categories and include human responses. This benchmark also allows us to compare models with humans in object recognition in challenging OOD scenarios. The collection includes sketches, edge-filtered images, silhouettes, images with texture-shape cue conflicts, and stylized images with textures replaced by painting styles. Additionally, twelve datasets involve parametric image degradation and varying factors such as noise and blur. Images within these OOD datasets were sourced from various datasets \citep{wang2019learning, geirhos2019imagenet, wichmann2017methods,geirhos2018generalisation,geirhos2019imagenet}. OOD accuracy is defined as the mean accuracy of a model across these 17 datasets, providing a comprehensive measure of OOD performance. Shape bias is the ratio of model accuracy on shapes to the sum of accuracy on shapes and textures in the cue-conflict dataset. 

\section{Results}
In this work, we study the characteristics of different models regarding human alignment based on their parameter size, the dataset they were trained on, and their learning methods. We also examine the relationship between human alignment metrics and the robustness of models.

\subsection{Impact of Scaling on Frequency Bandwidth}


\textbf{Model Scaling.} We conduct a comparative analysis of the frequency bandwidth of various models relative to humans by increasing the model sizes, irrespective of the underlying architecture, learning objective, or data augmentation methodologies. Figure \ref{fig:model_scale} demonstrates that with the increase of model size (X-axis), there is a reduction in bandwidth (Y-axis), signifying a progression towards human-level performance. By extrapolating this trend line, we can predict that models with approximately 31 billion parameters have the potential to achieve a human-level one-octave bandwidth.


  \textbf{Data Scaling.} We examine the effect of data scaling and training on a larger number of categories on the model's bandwidth. Figure \ref{fig:data_scale} shows that models trained in ImageNet-22K, a data set with 22K labels, exhibit a bandwidth closer to that of humans. This highlights the importance of data scaling to achieve human-like capabilities in visual tasks.
  
 Additionally, In Table \ref{tab:tab:beit_clip}, we observe that models trained on the LAION-2B \cite{schuhmann2022laion} dataset initially show a smaller bandwidth. However, fine-tuning these models on ImageNet-1K increases the bandwidth, and further fine-tuning on ImageNet-22K (with 22K classes) before final tuning on a subset of ImageNet-22K significantly improves the bandwidth, bringing it even closer to human levels (see Table \ref{tab:tab:beit_clip}).

\subsection{Relationship of BW to OOD Accuracy}
We also examine different metrics (frequency bandwidth, shape bias) that more accurately predict OOD accuracy. Figure \ref{fig:BW_ShB} demonstrates that bandwidth is a superior predictor of OOD accuracy when considering all networks in the regression analysis. As the BW decreases (approach towards humans), OOD accuracy increases. This inverse relationship (with a strong negative correlation of $r=-0.4$) underscores the importance of bandwidth as a predictive metric for the generalization of OOD, compared to shape bias, which has a weaker positive correlation ($r=0.13$). 

\begin{figure*}[htbp]  
    \centering
    \includegraphics[width=0.8\textwidth]{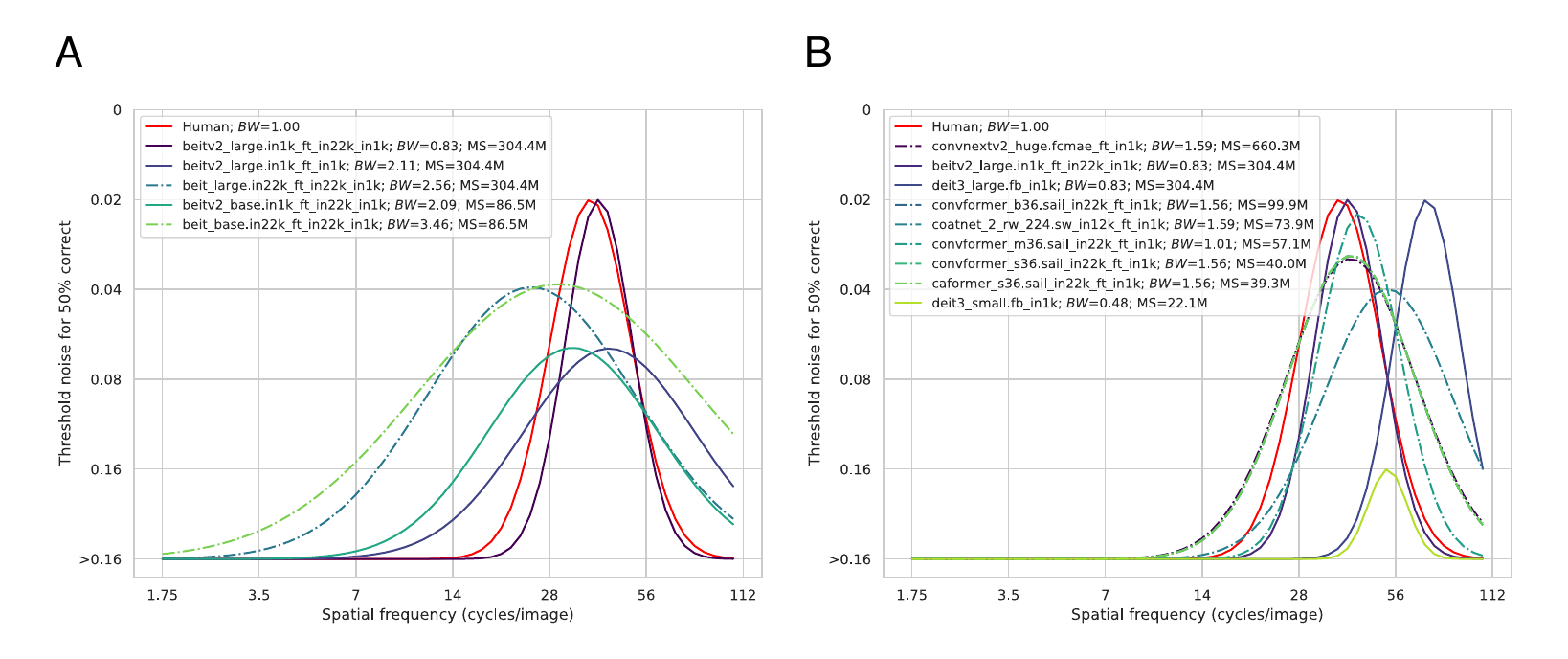}
\caption{Case studies. (A) Comparing Bandwidth of BEiT Models. (B) Human-like models. Many models exhibit human-like bandwidth, with a version of the BEiT-V2 model perfectly matching the human curve.}
    \label{fig:case_studies}
\end{figure*}

\subsection{Language Guidance leads to Human-like Bandwidth}
We investigated models that show the most human-like performance as case studies. In Figure \ref{fig:case_studies}, we show BEiT families with different setups. BEiTv2 \cite{Peng2022BEiTVM} uses a semantically rich visual tokenizer (distilling knowledge from multimodal pre-trained CLIP model) compared to the original BEiT \cite{bao2021beit}. Integration of \textbf{semantic tokenization}, \textbf{fine-tuning on ImageNet-22K with}, and \textbf{model size} all contribute to aligning the model with the human frequency bandwidth.

In Figure \ref{fig:case_studies}B, several models \textbf{outperform humans} or have bandwidths close to human levels, warranting further exploration. Models using both convolution and attention mechanisms \cite{vaswani2017attention}, such as \emph{CoAtNet} \cite{dai2021coatnet} and \emph{ConvFormer} \cite{yu2023metaformer} and \emph{DeiT-III} \cite{Touvron2022DeiTIR} with frequency-based data augmentation, and \emph{ConvNeXt-V2} \cite{woo2023convnext} with masked autoencoders and global response normalization show small bandwidths. More research is needed to understand these models, suggesting a direction for future studies.


\section{Discussion}

In this paper, our goal was to investigate various factors affecting the robustness of computer vision models and their alignment with human capabilities. We aimed to understand how model scaling, data scaling, semantic richness, data augmentation, large language model supervision, and multi-modality contribute to the performance of these models. 

Our study revealed several key findings. Firstly, increasing the number of parameters through model scaling brings models closer to human performance. Furthermore, we find scaling up the training data through data scaling results in decreased bandwidth. Moreover, providing more detailed information about the data helps models learn better representations which echo the findings of \cite{hong2023towards}. Furthermore, data augmentation with noise improves the robustness of models. Methods that use CLIP instead of supervised learning with labels, called large language model supervision, preserve more information and are robust against noise distortion. 
Finally, incorporating multiple modalities helps foundation models to learn semantics.

There are many illusions where human vision does not perceive the actual facts about an image \cite{anderson1997theory}, suggesting that humans are prone to mistakes. This observation might indicate that a model capable of surpassing human performance could achieve superhuman vision. Additionally, it might signify that humans utilize a wealth of contextual information to perceive images. For instance, in the checker shadow illusion, humans interpret two squares with the same shade of gray as differently coloured white and black squares. This phenomenon highlights the complex and often non-literal nature of human visual perception, which incorporates contextual cues and prior knowledge to construct a coherent understanding of visual stimuli.

These findings contribute to our understanding of the factors that influence the performance of computer vision models and provide insight into improving their alignment with human capabilities. 

\section{Conclusion}
Our results lead us to think that scaling foundation models might be the path to more robust machine learning models. However, several questions need to be answered:
1) While models are closing the gap with humans, what would be the next benchmark? Would new benchmarks such as OpenEQA \cite{majumdar2024openeqa} that evaluate models' capability performance in different aspects beyond object recognition be necessary?
2) In the future, can we simulate human vision with a foundation model? Can we use these models to cure and study the brain?

\section*{Impact Statement}

In practical applications where computer vision models are utilized as human assistants, these systems must mimic human behaviors and show robustness. Our work compares these systems by exploring their characteristics. We aim to contribute to the development of AI systems that prioritize safety, reliability, and ethical considerations in real-world scenarios.

        
\section*{Acknowledgements}

This work was funded by the Canada CIFAR AI Chair Program from the Canada Excellence Research Chairs (CERC) program and NSERC discovery grant RGPIN-2022-05122. We thank the Digital Research Alliance of Canada and Mila for providing computational resources.

\bibliography{ref}
\bibliographystyle{icml2024}

\newpage
\appendix
\onecolumn
\section{Checker shadow Illusion} \label{Checkershadow_Illusion}
\begin{figure*}[th]  
    \centering
    \includegraphics[width=0.75\textwidth]{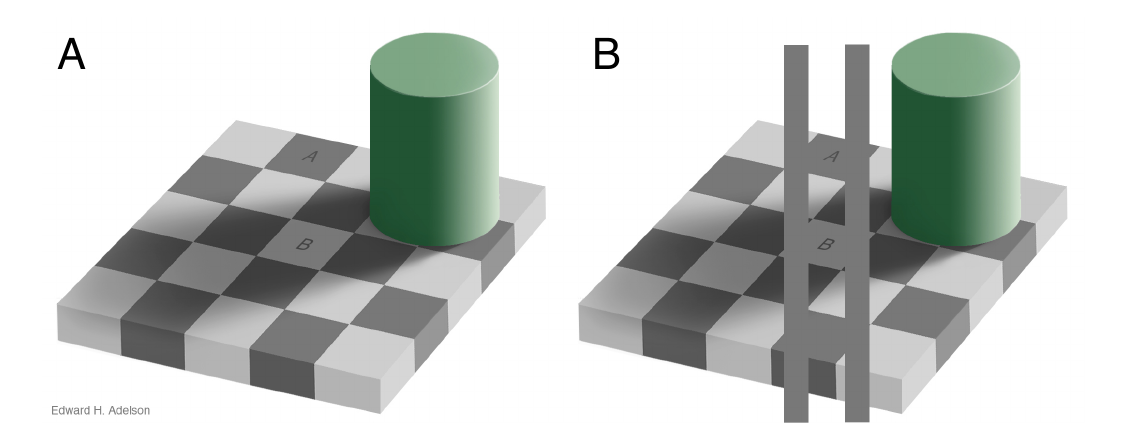}
    \caption{Checker Shadow Illusion (from \cite{adelson1995checker}). A) The squares marked A and B are actually the same shade of gray, yet they appear different due to the surrounding context. B) By connecting the squares marked A and B with vertical stripes of the same shade of gray, it becomes evident that both squares are indeed the same shade.}
    \label{fig:Checkershadow_Illusion}
\end{figure*}
The checkerboard shadow illusion shows how context affects how we perceive brightness and color. Due to the checkerboard pattern and shadows around them, two identically colored squares in this illusion appear to be different hues. In particular, one perceives a square in shadow as lighter (white), whereas one perceives the same square in well-lit areas as darker (black). This illusion draws attention to the intricate processes underlying human vision, whereby the brain interprets visual stimuli using contextual knowledge, frequently resulting in incorrect perceptions.

\section{Tables} \label{sec:table}

Table \ref{tab:tab:beit_clip} represents various setups of ViT-L/16 to elucidate why BEiT-V2 training achieves human-like behavior and its connection to other metrics such as OOD accuracy and shape bias. BEiT-V2 utilizes CLIP ViT-B/16 as a teacher for tokenization, resulting in superior accuracy and shape bias compared to BeiT and ViT-L/16.
Analysis reveals that both CLIP supervision and training on ImageNet-22K contribute to bandwidth, with ImageNet-22K having a more pronounced effect. In the final section of the table, we compare OpenClip, trained on Laion-2B and then fine-tuned on subsets of ImageNet-22K (ImageNet-12K) and ImageNet-1K, and only fine-tuned on ImageNet-1K. The results demonstrate that fine-tuning on ImageNet-1K alone adversely affects bandwidth. Moreover, increasing the number of labels improves shape bias, but the trend in OOD accuracy differs.

\input{tabel_models}

\section{Scaling Experiment Details}
\label{sec:Scaling_Experiment_details}
\begin{figure*}[tbh] 

  \centering
  \includegraphics[width=0.75\textwidth]{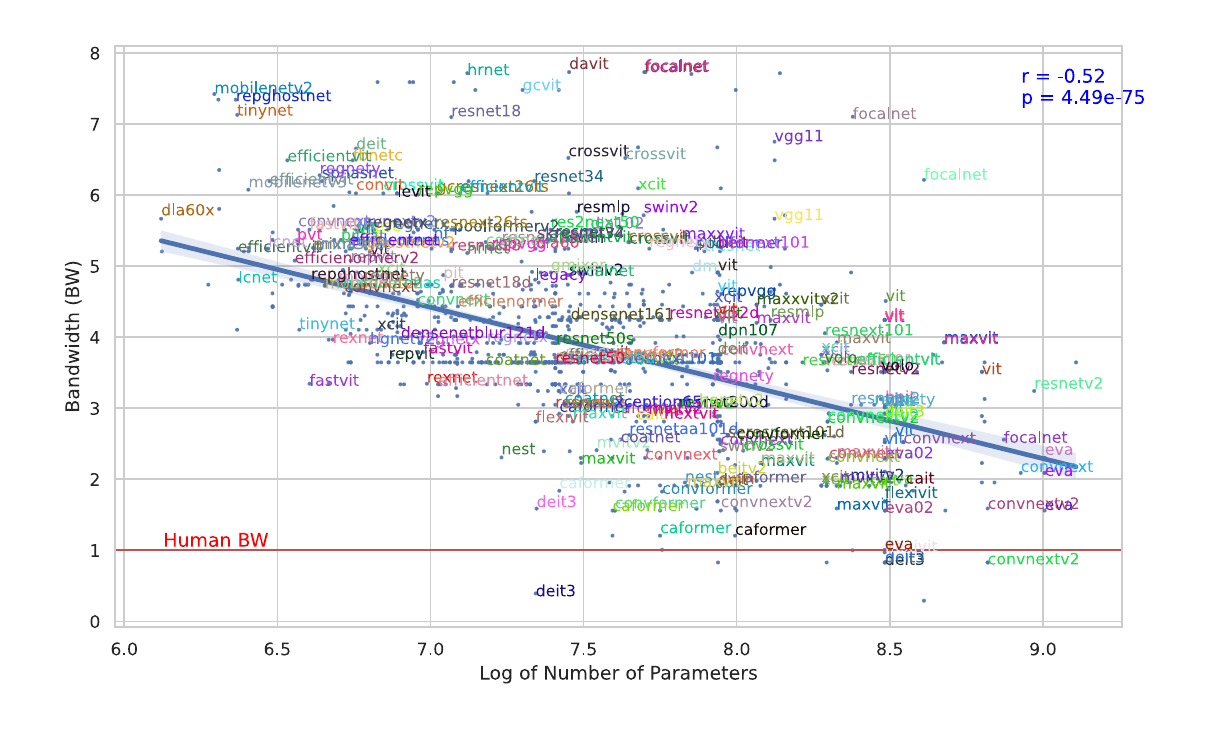}
\caption{Correlation between Bandwidth (BW) and Model Size in Logarithmic Scale. The regression line represents that as the model size scales, the bandwidth decreases, converging towards human levels. Each dot corresponds to a model, and some of the model names are shown.}
  \label{fig:model_scale_appendix}
\end{figure*}


\end{document}

%% file: tabel_models.tex

\begin{table*}[htbp] 
\centering
\begin{threeparttable}[b]
\caption{Factors Contributing to Model Alignment with Human Bandwidth. The best performing configurations are highlighted in green and the second best in yellow. The results indicate that using CLIP ViT-B/16 for tokenization and training on ImageNet-22K enhances the performance of BEiT-V2 ViT-L/16, bringing it closer to the one-octave bandwidth characteristic of human vision.}
\label{tab:tab:beit_clip}
\begin{tabular}{@{}lccccrrrr@{}}
\toprule
\cellcolor[HTML]{FFFFFF}Model &
  \cellcolor[HTML]{FFFFFF}Z-Shot &
  \cellcolor[HTML]{FFFFFF}CLIP &
  \cellcolor[HTML]{FFFFFF}IN-1k &
  \cellcolor[HTML]{FFFFFF}IN-22k &
  \multicolumn{1}{l}{\cellcolor[HTML]{FFFFFF}BW} &
  \multicolumn{1}{l}{OOD} &
  \multicolumn{1}{l}{Shape Bias} &
  \\ \midrule
\cellcolor[HTML]{FFFFFF}Humans &
  \multicolumn{1}{l}{} &
  \multicolumn{1}{l}{} &
  \multicolumn{1}{l}{\cellcolor[HTML]{FFFFFF}{\color[HTML]{4D5156} }} &
  \multicolumn{1}{l}{\cellcolor[HTML]{FFFFFF}{\color[HTML]{4D5156} }} &
  \cellcolor[HTML]{FFFFFF}1.0000 &
  \cellcolor[HTML]{FFFFFF}0.7304 &
  \cellcolor[HTML]{FFFFFF}0.9600 &
  \\ \midrule
  \rowcolor[HTML]{FFFFFF} 
CLIP ViT-B/16 \tnote{1} &
  {\color[HTML]{4D5156} \checkmark} &
  {\color[HTML]{4D5156} \checkmark} &
  {\color[HTML]{4D5156} ×} &
  {\color[HTML]{4D5156} ×} &
  2.7556 &
  0.6950 &
  0.4731 &
  \\ \midrule
\rowcolor[HTML]{FFFFFF} 
Original ViT-L/16 &
  {\color[HTML]{4D5156} ×} &
  {\color[HTML]{4D5156} ×} &
  {\color[HTML]{4D5156} \checkmark} &
  {\color[HTML]{4D5156} \checkmark} &
  3.5121 &
  0.7200 &
  \cellcolor[HTML]{FFD666}0.5381 &
  \\

\cellcolor[HTML]{FFFFFF}BEiT ViT-L/16 &
  × &
  \cellcolor[HTML]{FFFFFF}{\color[HTML]{4D5156} ×} &
  \cellcolor[HTML]{FFFFFF}{\color[HTML]{4D5156} \checkmark} &
  \cellcolor[HTML]{FFFFFF}{\color[HTML]{4D5156} \checkmark} &
  \cellcolor[HTML]{FFFFFF}2.5577 &
  \cellcolor[HTML]{FFFFFF}0.4898 &
  \cellcolor[HTML]{FFFFFF}0.4411 &
  \\
\cellcolor[HTML]{FFFFFF}BEiT-V2 ViT-L/16 &
  × &
  \cellcolor[HTML]{FFFFFF}{\color[HTML]{4D5156} \checkmark} &
  \cellcolor[HTML]{FFFFFF}{\color[HTML]{4D5156} \checkmark} &
  \cellcolor[HTML]{FFFFFF}{\color[HTML]{4D5156} ×} &
  \cellcolor[HTML]{FFD666}2.1084 &
  \cellcolor[HTML]{FFD666}0.7332 &
  \cellcolor[HTML]{FFFFFF}0.5364 &
  \\
\rowcolor[HTML]{FFFFFF} 
BEiT-V2 ViT-L/16 &
  {\color[HTML]{4D5156} ×} &
  {\color[HTML]{4D5156} \checkmark} &
  {\color[HTML]{4D5156} \checkmark} &
  {\color[HTML]{4D5156} \checkmark} &
  \cellcolor[HTML]{B6E1CD}0.8285 &
  \cellcolor[HTML]{B6E1CD}0.7560 &
  \cellcolor[HTML]{B6E1CD}0.5610 &
  \\ \midrule
\cellcolor[HTML]{FFFFFF}OpenCLIP ViT-L/14 &
  \cellcolor[HTML]{FFFFFF}{\color[HTML]{4D5156} \checkmark} &
  \checkmark &
  × &
  × &
  \cellcolor[HTML]{FFFFFF}2.8895 &
  \cellcolor[HTML]{FFFFFF}0.6931 &
  \cellcolor[HTML]{FFFFFF}0.5665 &
  \\
\cellcolor[HTML]{FFFFFF}OpenCLIP ViT-L/14 ft-IN1k &
  × &
  \checkmark &
  \checkmark &
  × &
  \cellcolor[HTML]{FFFFFF}3.7526 &
  \cellcolor[HTML]{FFFFFF}0.7184 &
  \cellcolor[HTML]{FFFFFF}0.4738 &
  \\ 
\cellcolor[HTML]{FFFFFF}OpenCLIP ViT-L/14  ft-IN12k-IN1k &
  × &
  \checkmark &
  \checkmark &
  \checkmark &
  \cellcolor[HTML]{FFFFFF}2.5012 &
  \cellcolor[HTML]{FFFFFF}0.7401 &
  \cellcolor[HTML]{FFFFFF}0.5121 &
  \\ \bottomrule
\end{tabular}
     \begin{tablenotes}
       \item [1] Teacher for BEiT-V2 tokenizer.
     \end{tablenotes}
  \end{threeparttable}
\end{table*}

%% file: main.bbl
\begin{thebibliography}{36}
\providecommand{\natexlab}[1]{#1}
\providecommand{\url}[1]{\texttt{#1}}
\expandafter\ifx\csname urlstyle\endcsname\relax
  \providecommand{\doi}[1]{doi: #1}\else
  \providecommand{\doi}{doi: \begingroup \urlstyle{rm}\Url}\fi

\bibitem[Adelson(1995)]{adelson1995checker}
Adelson, E.
\newblock Checker-shadow illusion. retrieved august 27 2018, 1995.

\bibitem[Anderson(1997)]{anderson1997theory}
Anderson, B.~L.
\newblock A theory of illusory lightness and transparency in monocular and binocular images: The role of contour junctions.
\newblock \emph{Perception}, 26\penalty0 (4):\penalty0 419--453, 1997.

\bibitem[Bao et~al.(2021)Bao, Dong, Piao, and Wei]{bao2021beit}
Bao, H., Dong, L., Piao, S., and Wei, F.
\newblock Beit: Bert pre-training of image transformers.
\newblock \emph{arXiv preprint arXiv:2106.08254}, 2021.

\bibitem[Campbell \& Robson(1968)Campbell and Robson]{campbell1968application}
Campbell, F.~W. and Robson, J.~G.
\newblock Application of fourier analysis to the visibility of gratings.
\newblock \emph{The Journal of physiology}, 197\penalty0 (3):\penalty0 551, 1968.

\bibitem[Cherti et~al.(2023)Cherti, Beaumont, Wightman, Wortsman, Ilharco, Gordon, Schuhmann, Schmidt, and Jitsev]{cherti2023reproducible}
Cherti, M., Beaumont, R., Wightman, R., Wortsman, M., Ilharco, G., Gordon, C., Schuhmann, C., Schmidt, L., and Jitsev, J.
\newblock Reproducible scaling laws for contrastive language-image learning.
\newblock In \emph{Proceedings of the IEEE/CVF Conference on Computer Vision and Pattern Recognition}, pp.\  2818--2829, 2023.

\bibitem[Dai et~al.(2021)Dai, Liu, Le, and Tan]{dai2021coatnet}
Dai, Z., Liu, H., Le, Q.~V., and Tan, M.
\newblock Coatnet: Marrying convolution and attention for all data sizes.
\newblock \emph{Advances in neural information processing systems}, 34:\penalty0 3965--3977, 2021.

\bibitem[Du et~al.(2020)Du, Zhen, Shao, and Snoek]{du2020metanorm}
Du, Y., Zhen, X., Shao, L., and Snoek, C.~G.
\newblock Metanorm: Learning to normalize few-shot batches across domains.
\newblock In \emph{International Conference on Learning Representations}, 2020.

\bibitem[Fel et~al.(2022)Fel, Rodriguez~Rodriguez, Linsley, and Serre]{fel2022harmonizing}
Fel, T., Rodriguez~Rodriguez, I.~F., Linsley, D., and Serre, T.
\newblock Harmonizing the object recognition strategies of deep neural networks with humans.
\newblock \emph{Advances in neural information processing systems}, 35:\penalty0 9432--9446, 2022.

\bibitem[Geirhos et~al.(2018{\natexlab{a}})Geirhos, Rubisch, Michaelis, Bethge, Wichmann, and Brendel]{geirhos2018imagenet}
Geirhos, R., Rubisch, P., Michaelis, C., Bethge, M., Wichmann, F.~A., and Brendel, W.
\newblock Imagenet-trained cnns are biased towards texture; increasing shape bias improves accuracy and robustness.
\newblock \emph{arXiv preprint arXiv:1811.12231}, 2018{\natexlab{a}}.

\bibitem[Geirhos et~al.(2018{\natexlab{b}})Geirhos, Temme, Rauber, Sch{\"u}tt, Bethge, and Wichmann]{geirhos2018generalisation}
Geirhos, R., Temme, C.~R., Rauber, J., Sch{\"u}tt, H.~H., Bethge, M., and Wichmann, F.~A.
\newblock Generalisation in humans and deep neural networks.
\newblock \emph{Advances in neural information processing systems}, 31, 2018{\natexlab{b}}.

\bibitem[Geirhos et~al.(2019{\natexlab{a}})Geirhos, Rubisch, Michaelis, Bethge, Wichmann, and Brendel]{geirhos2018imagenettrained}
Geirhos, R., Rubisch, P., Michaelis, C., Bethge, M., Wichmann, F.~A., and Brendel, W.
\newblock Imagenet-trained {CNN}s are biased towards texture; increasing shape bias improves accuracy and robustness.
\newblock In \emph{International Conference on Learning Representations}, 2019{\natexlab{a}}.
\newblock URL \url{https://openreview.net/forum?id=Bygh9j09KX}.

\bibitem[Geirhos et~al.(2019{\natexlab{b}})Geirhos, Rubisch, Michaelis, Bethge, Wichmann, and Brendel]{geirhos2019imagenet}
Geirhos, R., Rubisch, P., Michaelis, C., Bethge, M., Wichmann, F.~A., and Brendel, W.
\newblock Imagenet-trained cnns are biased towards texture; increasing shape bias improves accuracy and robustness.
\newblock In \emph{International Conference on Learning Representations}, 2019{\natexlab{b}}.

\bibitem[Geirhos et~al.(2021)Geirhos, Narayanappa, Mitzkus, Thieringer, Bethge, Wichmann, and Brendel]{geirhos2021partial}
Geirhos, R., Narayanappa, K., Mitzkus, B., Thieringer, T., Bethge, M., Wichmann, F.~A., and Brendel, W.
\newblock Partial success in closing the gap between human and machine vision.
\newblock \emph{Advances in Neural Information Processing Systems}, 34:\penalty0 23885--23899, 2021.

\bibitem[Gulrajani \& Lopez-Paz(2020)Gulrajani and Lopez-Paz]{gulrajani2020search}
Gulrajani, I. and Lopez-Paz, D.
\newblock In search of lost domain generalization.
\newblock \emph{arXiv preprint arXiv:2007.01434}, 2020.

\bibitem[Hong et~al.(2023)Hong, Cui, Fuxman, Chan, and Luo]{hong2023towards}
Hong, G.~Z., Cui, Y., Fuxman, A., Chan, S.~H., and Luo, E.
\newblock Towards understanding the effect of pretraining label granularity.
\newblock \emph{arXiv preprint arXiv:2303.16887}, 2023.

\bibitem[Ilharco et~al.(2021)Ilharco, Wortsman, Wightman, Gordon, Carlini, Taori, Dave, Shankar, Namkoong, Miller, Hajishirzi, Farhadi, and Schmidt]{ilharco_gabriel_2021_5143773}
Ilharco, G., Wortsman, M., Wightman, R., Gordon, C., Carlini, N., Taori, R., Dave, A., Shankar, V., Namkoong, H., Miller, J., Hajishirzi, H., Farhadi, A., and Schmidt, L.
\newblock Openclip, 2021.
\newblock URL \url{https://doi.org/10.5281/zenodo.5143773}.

\bibitem[Koh et~al.(2021)Koh, Sagawa, Marklund, Xie, Zhang, Balsubramani, Hu, Yasunaga, Phillips, Gao, et~al.]{koh2021wilds}
Koh, P.~W., Sagawa, S., Marklund, H., Xie, S.~M., Zhang, M., Balsubramani, A., Hu, W., Yasunaga, M., Phillips, R.~L., Gao, I., et~al.
\newblock Wilds: A benchmark of in-the-wild distribution shifts.
\newblock In \emph{International conference on machine learning}, pp.\  5637--5664. PMLR, 2021.

\bibitem[Li et~al.(2018)Li, Yang, Song, and Hospedales]{li2018learning}
Li, D., Yang, Y., Song, Y.-Z., and Hospedales, T.
\newblock Learning to generalize: Meta-learning for domain generalization.
\newblock In \emph{Proceedings of the AAAI conference on artificial intelligence}, volume~32, 2018.

\bibitem[Majaj et~al.(2002)Majaj, Pelli, Kurshan, and Palomares]{majaj2002role}
Majaj, N.~J., Pelli, D.~G., Kurshan, P., and Palomares, M.
\newblock The role of spatial frequency channels in letter identification.
\newblock \emph{Vision research}, 42\penalty0 (9):\penalty0 1165--1184, 2002.

\bibitem[Majumdar et~al.(2024)Majumdar, Ajay, Zhang, Putta, Yenamandra, Henaff, Silwal, Mcvay, Maksymets, Arnaud, et~al.]{majumdar2024openeqa}
Majumdar, A., Ajay, A., Zhang, X., Putta, P., Yenamandra, S., Henaff, M., Silwal, S., Mcvay, P., Maksymets, O., Arnaud, S., et~al.
\newblock Openeqa: Embodied question answering in the era of foundation models.
\newblock In \emph{2nd Workshop on Mobile Manipulation and Embodied Intelligence at ICRA 2024}, 2024.

\bibitem[Peng et~al.(2022)Peng, Dong, Bao, Ye, and Wei]{Peng2022BEiTVM}
Peng, Z., Dong, L., Bao, H., Ye, Q., and Wei, F.
\newblock Beit v2: Masked image modeling with vector-quantized visual tokenizers.
\newblock \emph{ArXiv}, abs/2208.06366, 2022.
\newblock URL \url{https://api.semanticscholar.org/CorpusID:251554649}.

\bibitem[Radford et~al.(2021)Radford, Kim, Hallacy, Ramesh, Goh, Agarwal, Sastry, Askell, Mishkin, Clark, Krueger, and Sutskever]{pmlr-v139-radford21a}
Radford, A., Kim, J.~W., Hallacy, C., Ramesh, A., Goh, G., Agarwal, S., Sastry, G., Askell, A., Mishkin, P., Clark, J., Krueger, G., and Sutskever, I.
\newblock Learning transferable visual models from natural language supervision.
\newblock In Meila, M. and Zhang, T. (eds.), \emph{Proceedings of the 38th International Conference on Machine Learning}, volume 139 of \emph{Proceedings of Machine Learning Research}, pp.\  8748--8763. PMLR, 18--24 Jul 2021.
\newblock URL \url{https://proceedings.mlr.press/v139/radford21a.html}.

\bibitem[Sagawa et~al.(2019)Sagawa, Koh, Hashimoto, and Liang]{sagawa2019distributionally}
Sagawa, S., Koh, P.~W., Hashimoto, T.~B., and Liang, P.
\newblock Distributionally robust neural networks for group shifts: On the importance of regularization for worst-case generalization.
\newblock \emph{arXiv preprint arXiv:1911.08731}, 2019.

\bibitem[Schuhmann et~al.(2022)Schuhmann, Beaumont, Vencu, Gordon, Wightman, Cherti, Coombes, Katta, Mullis, Wortsman, et~al.]{schuhmann2022laion}
Schuhmann, C., Beaumont, R., Vencu, R., Gordon, C., Wightman, R., Cherti, M., Coombes, T., Katta, A., Mullis, C., Wortsman, M., et~al.
\newblock Laion-5b: An open large-scale dataset for training next generation image-text models.
\newblock \emph{Advances in Neural Information Processing Systems}, 35:\penalty0 25278--25294, 2022.

\bibitem[Shi et~al.(2021)Shi, Seely, Torr, Siddharth, Hannun, Usunier, and Synnaeve]{shi2021gradient}
Shi, Y., Seely, J., Torr, P.~H., Siddharth, N., Hannun, A., Usunier, N., and Synnaeve, G.
\newblock Gradient matching for domain generalization.
\newblock \emph{arXiv preprint arXiv:2104.09937}, 2021.

\bibitem[Solomon \& Pelli(1994)Solomon and Pelli]{solomon1994visual}
Solomon, J.~A. and Pelli, D.~G.
\newblock The visual filter mediating letter identification.
\newblock \emph{Nature}, 369\penalty0 (6479):\penalty0 395--397, 1994.

\bibitem[Subramanian et~al.(2024)Subramanian, Sizikova, Majaj, and Pelli]{subramanian2024spatial}
Subramanian, A., Sizikova, E., Majaj, N., and Pelli, D.
\newblock Spatial-frequency channels, shape bias, and adversarial robustness.
\newblock \emph{Advances in Neural Information Processing Systems}, 36, 2024.

\bibitem[Sun \& Saenko(2016)Sun and Saenko]{sun2016deep}
Sun, B. and Saenko, K.
\newblock Deep coral: Correlation alignment for deep domain adaptation.
\newblock In \emph{Computer Vision--ECCV 2016 Workshops: Amsterdam, The Netherlands, October 8-10 and 15-16, 2016, Proceedings, Part III 14}, pp.\  443--450. Springer, 2016.

\bibitem[Touvron et~al.(2022)Touvron, Cord, and Jegou]{Touvron2022DeiTIR}
Touvron, H., Cord, M., and Jegou, H.
\newblock Deit iii: Revenge of the vit.
\newblock \emph{arXiv preprint arXiv:2204.07118}, 2022.

\bibitem[Vaswani et~al.(2017)Vaswani, Shazeer, Parmar, Uszkoreit, Jones, Gomez, Kaiser, and Polosukhin]{vaswani2017attention}
Vaswani, A., Shazeer, N., Parmar, N., Uszkoreit, J., Jones, L., Gomez, A.~N., Kaiser, {\L}., and Polosukhin, I.
\newblock Attention is all you need.
\newblock \emph{Advances in neural information processing systems}, 30, 2017.

\bibitem[Wang et~al.(2019)Wang, Ge, Lipton, and Xing]{wang2019learning}
Wang, H., Ge, S., Lipton, Z., and Xing, E.~P.
\newblock Learning robust global representations by penalizing local predictive power.
\newblock \emph{Advances in Neural Information Processing Systems}, 32, 2019.

\bibitem[Wichmann et~al.(2017)Wichmann, Janssen, Geirhos, Aguilar, Sch{\"u}tt, Maertens, and Bethge]{wichmann2017methods}
Wichmann, F.~A., Janssen, D.~H., Geirhos, R., Aguilar, G., Sch{\"u}tt, H.~H., Maertens, M., and Bethge, M.
\newblock Methods and measurements to compare men against machines.
\newblock \emph{Electronic Imaging}, 29:\penalty0 36--45, 2017.

\bibitem[Wightman(2019)]{rw2019timm}
Wightman, R.
\newblock Pytorch image models.
\newblock \url{https://github.com/rwightman/pytorch-image-models}, 2019.

\bibitem[Woo et~al.(2023)Woo, Debnath, Hu, Chen, Liu, Kweon, and Xie]{woo2023convnext}
Woo, S., Debnath, S., Hu, R., Chen, X., Liu, Z., Kweon, I.~S., and Xie, S.
\newblock Convnext v2: Co-designing and scaling convnets with masked autoencoders.
\newblock In \emph{Proceedings of the IEEE/CVF Conference on Computer Vision and Pattern Recognition}, pp.\  16133--16142, 2023.

\bibitem[Yu et~al.(2023)Yu, Si, Zhou, Luo, Zhou, Feng, Yan, and Wang]{yu2023metaformer}
Yu, W., Si, C., Zhou, P., Luo, M., Zhou, Y., Feng, J., Yan, S., and Wang, X.
\newblock Metaformer baselines for vision.
\newblock \emph{IEEE Transactions on Pattern Analysis and Machine Intelligence}, 2023.

\bibitem[Zhang et~al.(2022)Zhang, Zhou, Xu, Cui, Shen, and Liu]{zhang2022towards}
Zhang, X., Zhou, L., Xu, R., Cui, P., Shen, Z., and Liu, H.
\newblock Towards unsupervised domain generalization.
\newblock In \emph{Proceedings of the IEEE/CVF Conference on Computer Vision and Pattern Recognition}, pp.\  4910--4920, 2022.

\end{thebibliography}
